# Manage risks in complex engagements by leveraging organization-wide knowledge using Machine Learning

Hari Prasad, Akhil Goyal, Shivram Ramasubramanian

## Abstract

One of the ways for organizations to continuously get better at executing projects is to learn from their past experience. In large organizations, the different accounts and business units often work in silos and tapping the rich knowledge base across the organization is easier said than done. With easy access to the collective experience spread across the organization, project teams and business leaders can proactively anticipate and manage risks in new engagements. Early discovery and timely management of risks is key to success in today's complex engagements. In this paper, the authors describe a Machine Learning based solution deployed with MLOps principles to solve this problem in an efficient manner.

***Disclaimer:*** *The views expressed in this paper belong to the content creators, and not Cognizant, its affiliates, or employees.*

## Introduction

For project-centric organizations, cost-effective, differentiated delivery is key to success. Early understanding of risks and mitigations play a crucial role in achieving this. Our discussions with project managers and business leaders revealed a need to learn from the experience in similar projects, understand the risks they faced and plan to mitigate such risks in advance.

Such learning from the experience of similar projects executed across the enterprise results in significant business benefits.

- Early discovery of risks results in proactive risk mitigation, cost savings, enhanced customer satisfaction and increased revenue-generation opportunities.
- Collaboration between teams from similar projects helps in sharing of ideas and best practices to improve delivery quality and create a culture of knowledge-sharing.

Manually maintained rule-based methods to identify similar projects involve using multiple, restrictive, subspace search rules. Rules needs to be continuously managed and constantly updated. This approach has serious limitations.

- *Inability to do contextual text comparison*: It becomes an arduous task to define and maintain scalable rules to search similar terms, e.g., similar tools and technologies. It is almost impossible for the manual rules to scale and pick contextually similar risks.
- *Poor User Experience*: Using filters defined by manual rules results in a very restrictive subspace search, resulting in no results beyond a point. Users typically expect auto-populated results, rather than a filtering approach.

Hence a scalable, enterprise level, Machine Learning (ML) based solution is required to overcome these limitations.

# Solution

Our solution comprises of two components as outlined in Figure 1.

- **Project similarity**: This component identifies similar projects across the organization.
- **Risk similarity**: This component then maps the risks tracked in such similar projects to contextually similar risks from a set of curated risks.

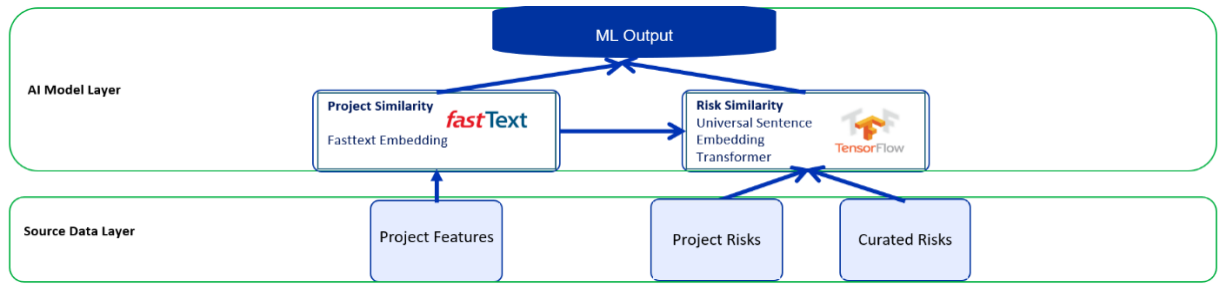

*Figure 1: Risk Discovery from Similar Projects Solution Architecture*

## Project similarity

There is no labeled data available that identifies similar projects; hence this needs to be an unsupervised ML solution. We envisioned project information as a collection of all important text that describes the project. This way of envisioning the project information is highly scalable as additional information that becomes available in future can be easily added as text without any change in architecture.

Significant expressions related to the project are extracted from the text using key phrase extraction algorithm [1]. The fastText [2] embedding is used instead of word2vec [3] due to its ability to produce rich word embeddings at sub-word level and its ability to handle minor misspellings. Arc cosine similarity is used to get similar projects instead of cosine similarity to avoid the limitation of flattening at the extreme ends of spectrum, making similar projects indistinguishable at high similarities.

## Risk discovery from similar projects

Risk discovery from similar projects involves discovering risks from the curated set that are contextually similar to the risks tracked in similar projects.

### Risk and mitigation curation

The risks tracked in similar projects often have reference to context and information that is very specific to the project. In order to make the risks and mitigation suggestions more useful to the users, it was decided to present these from a curated risk database instead of the raw risks from the similar projects. Commonly occurring risks and recommended mitigations were manually compiled by subject matter experts as a curated risk database, after analyzing historic risks and mitigations recorded in the enterprise risk platform.

### Contextual risk similarity

Universal Sentence Encoder [4], large transformer-based module trained and released by Google, Tensorflow [5] Hub has shown an excellent ability to understand the context of paragraphs and provide semantic similarity with high relevancy. It is used to get cosine similarity between the raw risks and the manually curated risks.

Highly similar, curated risks above a similarity threshold are extracted. This threshold was decided based on functional evaluation of a random set of raw risks and curated risks. The raw risks that do not have a matching curated risk above the threshold are periodically evaluated and accommodated by combination of the following approaches: 1. Creation of new curated risks 2. Usage of advanced pretrained modules 3. Siamese fine tuning of module using semantically similar, but low threshold inputs. The Appendix section of this paper has further details on the research conducted on Siamese fine tuning.

### Duplicate risk removal through semantic similarity

When a set of subject matter experts write curated risks in silos, same curated risks can be represented in different words, resulting in outputs with duplicate information to the end user. Hence prior to showing the risks to the end user, duplicate removal is done by applying the same risk similarity check on the interim output. As a result, only unique risks are presented to the user.

## Azure MLOps deployment

Microsoft Azure Machine Learning (ML) platform was chosen as the ML deployment platform to automate end to end flow of this solution using MLOps. Azure ML Pipelines are used to schedule and run the ML job frequently, connecting to Azure storage where project and risk data is stored. Registered models precompute similar projects and risks, to provide recommendations for a given project.

These models are deployed in scalable Azure Kubernetes clusters, and REST APIs are exposed to enterprise portals via secured Apigee gateway as shown in Figure 2.

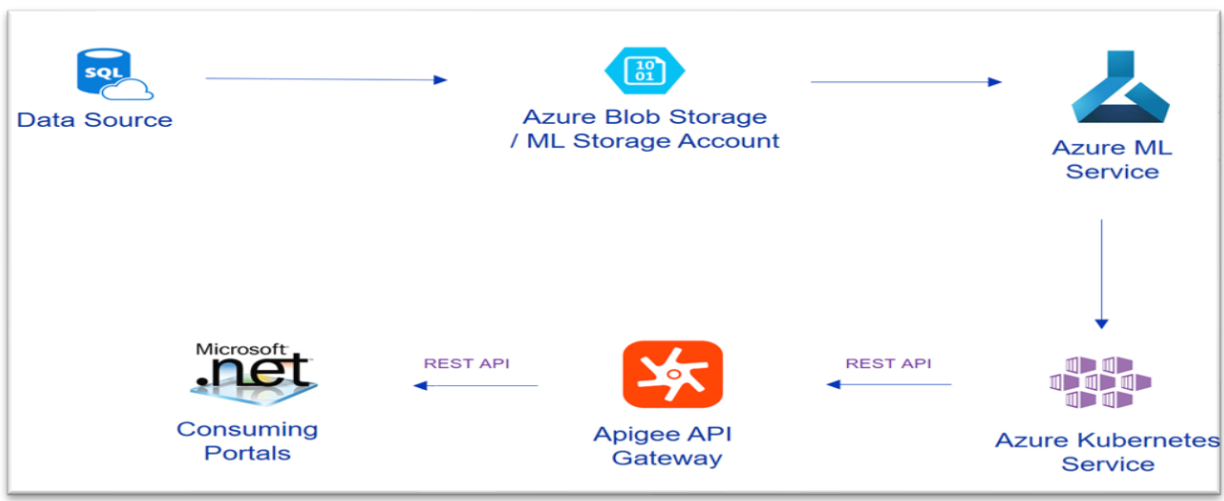

*Figure 2: Azure ML Deployment Solution Architecture*

## Business benefits

- **Enterprise knowledge discovery**: Integrated with the enterprise knowledge discovery portal, the solution presents learning from similar projects to the project owners. Here**,** collaboration options between similar project owners are provided through integration with the enterprise messaging and mailing platform where they can either chat or get connected over email.
- **Enterprise project management and risk discovery:** Risk suggestions from similar projects is integrated with the enterprise risk management platform. This enables project owners to discover relevant risks, assess recommendations to mitigate, import them in their project's risk register and act on these risks in a timely manner.

This solution can be used across all projects in the organizations. Following are a few real-life cases where the project teams benefitted from the solution.

- For a large electrical manufacturing client, the project team was working on an ecommerce platform Magento. Since there were limited projects on Magento in the repository, similar projects in Drupal were also identified. The solution was able to correlate the two related, competing technologies without being explicitly instructed to do so. Magento being a niche skill, relevant risks related to resource availability were highlighted.
- For a large UK telecom provider, we were running an ETL testing project on Ab Initio. In addition to listing similar projects doing Ab Initio testing, highly relevant risks related to inadequate ETL configuration in test environment leading to delay in testing and defect leakage were shown, along with suggestions to mitigate.
- For an Australian financial services client, the team was working on a development project with secure connectivity requirements. They were able to anticipate potential infrastructure challenges due to the COVID-enforced work from home setup upfront and planned ahead based on learning from similar projects.

## Further work

The manual risk and mitigation curation is an effort-intensive exercise. A hybrid approach to risk curation where an ML-led abstractive-summarization is reviewed by experts is in experimental stage. This is expected to assist the experts by substantially reducing their effort on risk curation.

Usage of advanced pretrained modules and Siamese fine tuning of the prebuilt module to uplift similarity scores of functionally similar, but low similarity score risks, is being experimented.

Work is also in progress to build a search functionality on curated risks which can provide the relevant risks based on search keywords, independent of the pipeline flow of this solution.

## Acknowledgements

We thank the functional experts for creating curated risks and mitigations, along with the iterative functional evaluation of the solution. We also would like to thank the experts from knowledge management and IT teams for integrating the solution with the relevant enterprise apps.

# Appendix: Siamese Fine Tuning

There will be a portion of base risks which will not find any matching curated risks above the similarity threshold when we use pretrained embedding modules without tuning. During the functional evaluation we found that some of these risks were functionally similar to already written curated risks and needed to be given higher similarity scores.

This led to the research related to Siamese fine tuning, where parallel corpus of the raw risks and corresponding curated risks are given to universal sentence embedding to fine-tune in a Siamese fine-tuning architecture to elevate the similarity scores. During this work a document improvement to Tensorflow Hub was suggested related to fine-tuning with a generic code. This change was accepted and published as a document improvement for the fine-tuning section of Tensorflow Hub document [6].

During fine-tuning experiments, it was observed that while fine-tuning increases the similarity of parallel corpus as per expectation, it also increased the similarity scores for others which were in low score region prior to fine-tuning. Sample parallel corpus cosine similarity results are presented in Table 1. The diagonal of the table represents parallel corpus similarity, while other values show intra parallel corpus similarity.

Using the Semantic Textual Similarity (STS) evaluation benchmark, out of the box module Pearson correlation coefficient is found to be at 0.78, with p value of $3.8e^{-285}$, whereas fine-tuned Pearson correlation coefficient is found to be at 0.75, with p value of $7.5e^{-254}$. This shows the drop in generalization post fine-tuning and the need for doing careful regularization during fine-tuning.

Further experiments are being conducted to ensure the results generalize well, using dropouts [7] and regularizations, before the fine-tuned module can replace the out of the box pretrained universal sentence embedding module.

|   | 0 | 1 | 2 | 3 | 4 |
|---|---|---|---|---|---|
| 0 | 0.460943 | 0.538673 | 0.595297 | 0.622900 | 0.665010 |
| 1 | 0.545224 | 0.843681 | 0.663601 | 0.537297 | 0.575361 |
| 2 | 0.522664 | 0.645566 | 0.927904 | 0.471392 | 0.531852 |
| 3 | 0.532698 | 0.592737 | 0.583253 | 0.667764 | 0.558558 |
| 4 | 0.465463 | 0.562364 | 0.644324 | 0.550721 | 0.584414 |

Post Fine Tuning Similarities

|   | 0 | 1 | 2 | 3 | 4 |
|---|---|---|---|---|---|
| 0 | 0.197151 | 0.187207 | 0.253937 | 0.306738 | 0.332011 |
| 1 | 0.293789 | 0.709240 | 0.336964 | 0.215192 | 0.266110 |
| 2 | 0.182323 | 0.270148 | 0.856555 | 0.047758 | 0.125578 |
| 3 | 0.167337 | 0.167952 | 0.168668 | 0.410686 | 0.175115 |
| 4 | 0.187545 | 0.169519 | 0.351853 | 0.230305 | 0.310929 |

Pre Fine Tuning Similarities

*Table 1: Siamese Fine Tuning Semantic Similarity Updates. Identical index in row & column indicates parallel corpus, making diagonal of similarity matrix as parallel corpus cosine similarity score.*